\documentclass{article}

\usepackage{times}
\usepackage{graphicx} 
\usepackage{subfigure} 

\usepackage{natbib}

\usepackage{url}
\usepackage{wrapfig}
\usepackage{booktabs}       
\usepackage{amsmath}
\usepackage[labelfont=bf,textfont={},font=small,justification=justified,indention=0cm]{caption}

\usepackage{algorithm}
\usepackage{algorithmic}

\usepackage{hyperref}

\usepackage[accepted]{icml2017}

\icmltitlerunning{Conditional Image Synthesis with Auxiliary Classifier GANs}

\begin{document} 

\twocolumn[
\icmltitle{Conditional Image Synthesis with Auxiliary Classifier GANs}



\icmlsetsymbol{equal}{*}

\begin{icmlauthorlist}
\icmlauthor{Augustus Odena}{to}
\icmlauthor{Christopher Olah}{to}
\icmlauthor{Jonathon Shlens}{to}
\end{icmlauthorlist}

\icmlaffiliation{to}{Google Brain}

\icmlcorrespondingauthor{Augustus Odena}{augustusodena@google.com}

\icmlkeywords{boring formatting information, machine learning, ICML}

\vskip 0.3in
]



\printAffiliationsAndNotice{}  

\begin{abstract} 
In this paper we introduce new methods for the improved
training of
generative adversarial networks (GANs) for image synthesis.
We construct a variant of GANs employing label conditioning
that results in $128\times 128$ resolution image samples exhibiting global coherence.
We expand on previous work for image
quality assessment to provide two new analyses for assessing the discriminability
and diversity of samples from class-conditional image synthesis models.
These analyses demonstrate that high resolution samples provide
class information not present in low resolution samples.
Across 1000 ImageNet classes, $128\times 128$ samples are more
than twice as discriminable as artificially resized $32\times 32$ samples.
In addition, 84.7\% of the classes have samples exhibiting diversity
comparable to real ImageNet data.
\end{abstract} 

\section{Introduction}

Characterizing the structure of natural images has been a rich research endeavor.
Natural images obey intrinsic invariances and exhibit multi-scale statistical
structures that have historically been difficult to quantify
\citep{simoncelli2001review}.
Recent advances in machine learning offer an opportunity to substantially
improve the quality of image models. Improved image models advance the
state-of-the-art in
image denoising \citep{DENOISING},
compression \citep{toderici},
in-painting \citep{PIXELRNN}, and
super-resolution \citep{SUPERRESOLUTION}.
Better models of natural images also improve performance in
semi-supervised learning tasks
\citep{SSVAE, CATGAN, SGAN, IMPROVEDTECHNIQUES}
and reinforcement learning problems \citep{EPISODIC}.

One method for understanding natural image statistics
is to build a system that synthesizes
images \textit{de novo}.
There are several promising approaches for building image synthesis models.
Variational autoencoders (VAEs) maximize a variational lower bound
on the log-likelihood of the training data \citep{VAES, VAES2}.
VAEs are straightforward to train but
introduce potentially restrictive assumptions about the approximate
posterior distribution (but see \cite{NORMALIZINGFLOWS, IAF}).
Autoregressive models dispense with latent variables and directly model the conditional
distribution over pixels \citep{PIXELRNN, PIXELCNN}.
These models produce convincing samples but are costly to sample from
and do not provide a latent representation.
Invertible density estimators transform latent variables directly using a series of
parameterized functions constrained to be invertible \citep{nvp}.
This technique allows for exact log-likelihood computation and exact inference, but the invertibility constraint is restrictive.

Generative adversarial networks (GANs) offer a distinct and promising approach
that focuses on a game-theoretic formulation for training an image synthesis model \citep{GANS}.
Recent work has shown that GANs can produce convincing image samples on datasets with low variability and low resolution \citep{LAPGAN,DCGAN}.
However, GANs struggle to generate globally coherent, high resolution samples - 
particularly from datasets with high variability. Moreover, a theoretical understanding
of GANs is an on-going research topic \citep{THEORY1, THEORY2}.

\begin{figure*}[!t]
\hspace{-0.0\textwidth} 
\includegraphics[width=0.99\textwidth]{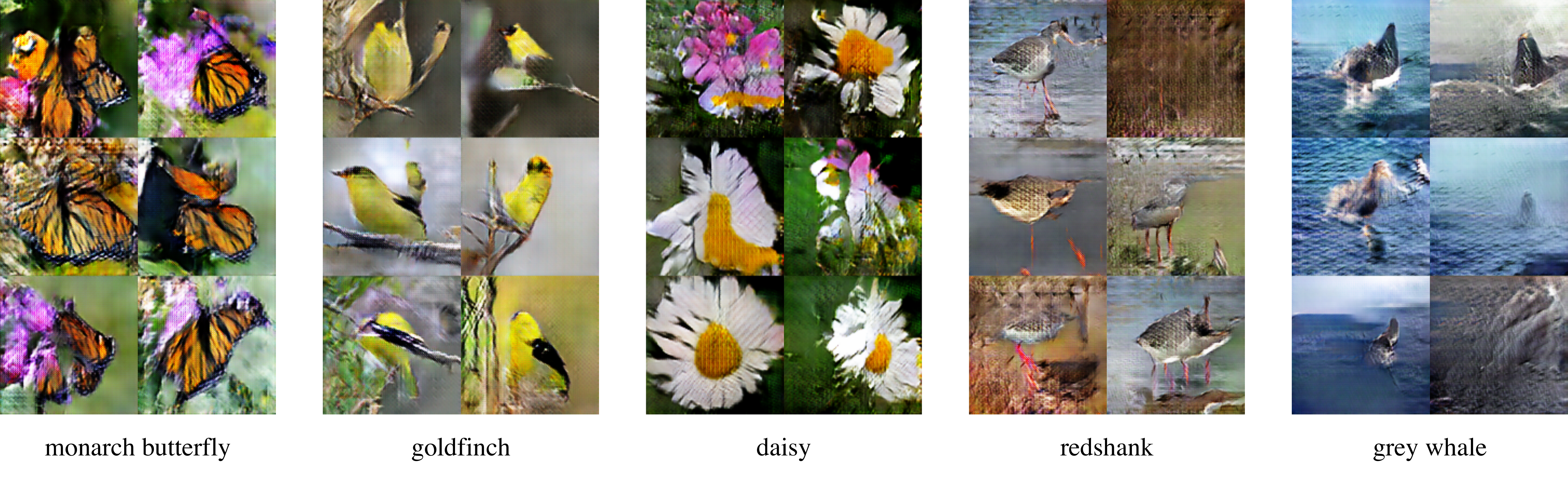}%
\vspace{-5pt}
\caption{
  $128\times 128$ resolution samples from 5 classes taken from an AC-GAN trained on the ImageNet dataset.
  Note that the classes shown have been selected to highlight the success of the model and are not representative. Samples from all ImageNet classes are linked later in the text.}
\label{fig:imagenetsamples}
\end{figure*}

In this work we demonstrate that that adding more structure to the GAN latent space
along with a specialized cost function results in higher quality samples.
We exhibit $128\times 128$ pixel samples from all classes of the ImageNet dataset \citep{IMAGENET} with increased global coherence (Figure \ref{fig:imagenetsamples}).
Importantly, we demonstrate quantitatively that our high resolution
samples are not just naive resizings of low resolution samples.
In particular, downsampling our $128 \times 128$ samples
to $32 \times 32$ leads to a 50\% decrease in visual discriminability.
We also introduce
a new metric for assessing the variability across image samples and employ this
metric to demonstrate that our synthesized images exhibit diversity comparable to
training data for a large fraction (84.7\%) of ImageNet classes.
In more detail, this work is the first to:

\begin{itemize}
\item Demonstrate an image synthesis model for all 1000 ImageNet classes at a 128x128 spatial resolution (or any spatial resolution - see Section \ref{section:ACGAN}).
\item Measure how much an image synthesis model actually uses its output resolution (Section \ref{section:resolution}).
\item Measure perceptual variability and 'collapsing' behavior in a GAN with a fast, easy-to-compute metric (Section \ref{section:diversity}).
\item Highlight that a high number of classes is what makes ImageNet synthesis difficult for GANs and provide an explicit solution (Section \ref{section:count}).
\item Demonstrate experimentally that GANs that perform well perceptually are not those that memorize a small number of examples (Section \ref{section:correlation}).
\item Achieve state of the art on the Inception score metric when trained on CIFAR-10 without using any of the techniques from \cite{IMPROVEDTECHNIQUES} (Section \ref{section:inceptionscore}).
\end{itemize}

\section{Background}

A generative adversarial network (GAN) consists of two neural networks trained in opposition to one another.
The generator $G$ takes as input a random noise vector $z$ and outputs an image $X_{fake} = G(z)$.
The discriminator $D$ receives as input either a training image or a synthesized
image from the generator and outputs a probability distribution $P(S ~|~ X)=D(X)$ over possible image sources.
The discriminator is trained to maximize the log-likelihood
it assigns to the correct source:
\begin{multline}
  \label{eqn:eqlabel}
L = E[\log P(S = real ~~|~ X_{real} )] + \\ E[\log P(S = fake ~|~ X_{fake})] $$
\end{multline}
The generator is trained to minimize the second term in Equation \ref{eqn:eqlabel}.

The basic GAN framework can be augmented using side information.
One strategy is to supply both the generator and discriminator
with class labels 
in order to produce class conditional samples \citep{CONDITIONAL}.
Class conditional synthesis can significantly improve the quality of
generated samples \citep{PIXELCNN}. 
Richer side information such as
image captions and bounding box localizations may improve
sample quality further \citep{REED2, REED1}.

Instead of feeding side information to the discriminator,
one can task the discriminator with reconstructing side information.
This is done by modifying the discriminator
to contain an auxiliary decoder network\footnote{
Alternatively, one can force the discriminator to work with the
joint distribution $(X,z)$
and train a separate inference network
 that computes $q(z|X)$ \citep{ALI, BIGAN}.
}
that outputs the class label for the training data \citep{SGAN, IMPROVEDTECHNIQUES}
or a subset of the latent variables from which the samples are generated \citep{INFOGAN}.
Forcing a model to perform additional tasks is known to 	
improve performance on the original task
(e.g. \cite{SequenceToSequence, INCEPTION, Ramsundar2015}).
In addition, an auxiliary decoder could leverage pre-trained discriminators
(e.g. image classifiers) for further improving the synthesized images \citep{nguyen2016}.
Motivated by these considerations, we introduce a model that combines
both strategies for leveraging side information.
That is, the model proposed below is class conditional,
but with an auxiliary decoder that is tasked with reconstructing class labels.

\section{AC-GANs} \label{section:ACGAN}

We propose a variant of the GAN architecture which we call an auxiliary classifier GAN
(or AC-GAN).
In the AC-GAN, every generated sample has a corresponding class label, $c \sim p_c$ in addition to the noise $z$.
$G$ uses both to generate images $X_{fake} = G(c,z)$.
The discriminator gives both a probability distribution over sources and a probability distribution over the class labels, $P(S ~|~ X),~ P(C ~|~ X) = D(X)$.
The objective function has two parts: the log-likelihood of the correct source, $L_S$,
and the log-likelihood of the correct class, $L_C$.

\begin{multline}
L_S = E[\log P(S = real ~~|~ X_{real} )] + \\ E[\log P(S = fake ~|~ X_{fake})]
\end{multline}
\begin{multline}
L_C = E[\log P(C = c ~~|~ X_{real} )] + \\ E[\log P(C = c ~|~ X_{fake})]
\end{multline}

$D$ is trained to maximize $L_S + L_C$ while $G$ is trained to maximize $L_C - L_S$.
AC-GANs learn a representation for $z$ that is independent of class label (e.g. \cite{SSVAE}).

Structurally, this model is not tremendously different from existing models.
However, this modification to the standard GAN formulation produces
excellent results and appears to stabilize training.
Moreover, we consider the AC-GAN model to be only part of the technical
contributions of this work, along with our proposed methods for
measuring the extent to which a model makes use of its given output resolution,
methods for measuring perceptual variability of samples from the model,
and a thorough experimental analyis of a generative model of images
that creates $128 \times 128$ samples from all 1000 ImageNet classes.

Early experiments demonstrated that increasing the number of
classes trained on while holding the model fixed decreased the quality
of the model outputs. The structure of the
AC-GAN model permits
separating large datasets into subsets by class
and training a generator and discriminator for each subset.
All ImageNet experiments are conducted using an ensemble of 100 AC-GANs, each trained on a 10-class split.

\section{Results}

We train several AC-GAN models on the ImageNet data set
\citep{IMAGENET}. Broadly speaking, the architecture of the generator $G$
is a series of `deconvolution' layers
that transform the noise
$z$ and class $c$ into an image
\citep{NNDECONV}. 
We train two variants of the model architecture for
generating images at $128\times 128$ and
$64\times 64$ spatial resolutions.
The discriminator $D$ is a
deep convolutional neural network with a Leaky ReLU nonlinearity \citep{Maas2013}.
As mentioned earlier, we find that reducing the variability introduced by
all 1000 classes of ImageNet significantly improves the quality of training.
We train 100 AC-GAN models -- each on images from just 10 classes
-- for 50000 mini-batches of size 100.

Evaluating the quality of image synthesis models is challenging due to 
the variety of probabilistic criteria \citep{TheisEtAl}
and the lack of a perceptually meaningful image similarity metric.
Nonetheless, in later sections we attempt to measure the quality of
the AC-GAN by building several \textit{ad-hoc} measures for
image sample discriminability and diversity.
Our hope is that this work might 
provide quantitative measures that may be used to aid training and subsequent
development of image synthesis models.

\subsection{Generating High Resolution Images Improves Discriminability} \label{section:resolution}

\begin{figure*}[t]
\includegraphics[width=0.69\textwidth]{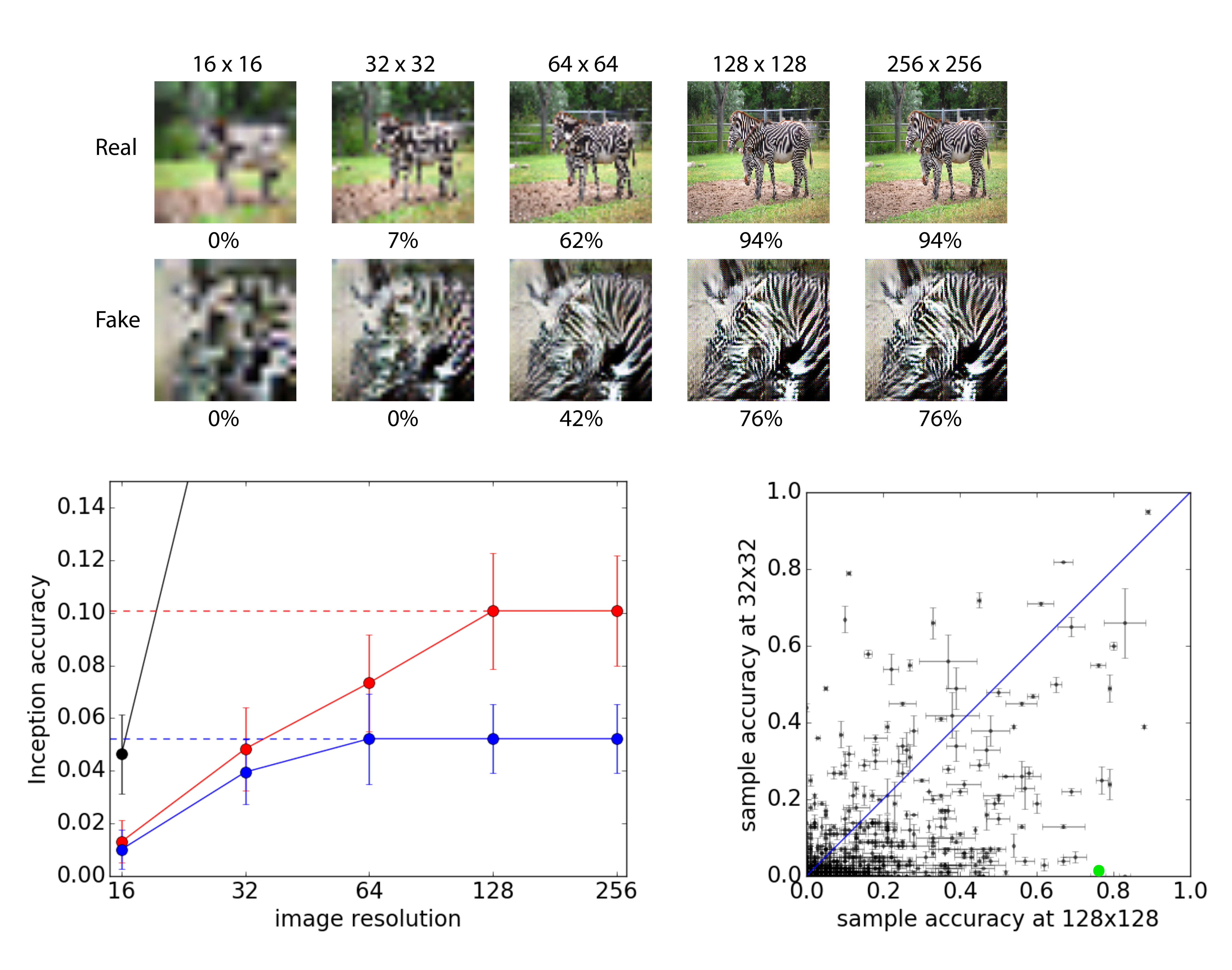}
\centering
\caption{Generating high resolution images improves discriminability. Top: Training data 
and synthesized images from the {\tt zebra} class resized to a lower spatial
resolution (indicated above) and subsequently artificially resized to the original resolution ($128 \times 128$ for the red and black lines; $64 \times 64$ for the blue line).
Inception accuracy is shown below the corresponding images.
Bottom Left:
Summary of accuracies across varying spatial resolutions for training data and image samples
from $64\times 64$ and $128\times 128$ models.
Error bar measures standard deviation across 10 subsets of images.
Dashed lines highlight the accuracy at the output spatial resolution of the model.
The training data (clipped) achieves accuracies of 24\%, 54\%, 81\% and 81\% at resolutions of
32, 64, 128, and 256 respectively.
Bottom Right:
Comparison of accuracy scores at $128\times 128$ and $32\times 32$ spatial resolutions ($x$ and $y$ axis, respectively).
Each point represents an ImageNet class.
84.4\% of the classes are below the line of equality.
The green dot corresponds to the {\tt zebra} class.
We also artificially resized $128\times 128$ and $64 \times 64$ images to $256\times 256$ as a sanity check
to demonstrate that simply increasing the number of pixels will not increase discriminability.}

\label{fig:inception_score_plot}
\end{figure*}

Building a class-conditional image synthesis model
necessitates measuring the extent to which synthesized images
appear to belong to the intended class.
In particular, we would like to know that a high resolution sample
is not just a naive resizing of a low resolution sample.
Consider a simple experiment: pretend there exists a model that synthesizes
$32\times 32$ images.
One can trivially increase the resolution of synthesized images
by performing bilinear interpolation.
This would yield higher resolution images, but these images would just be blurry
versions of the low resolution images that are not discriminable.
Hence, the goal of an image synthesis model is not simply to produce high resolution images,
but to produce high resolution images that are more discriminable than low resolution images.

To measure discriminability, we feed synthesized images to a pre-trained Inception network \citep{szegedy2015rethinking} and
report the fraction of the samples for which the Inception network assigned the correct
label\footnote{
  One could also use the Inception score \citep{IMPROVEDTECHNIQUES},
  but our method has several advantages:
  accuracy figures are easier to interpret than exponentiated KL-divergences;
  accuracy may be assessed for individual classes; accuracy measures whether
  a class-conditional model generated samples from the intended class.
  To compute the Inception accuracy, we modified a version of Inception-v3
  supplied in \texttt{https://github.com/openai/improved-gan/}.
}.
We calculate this accuracy measure on a series of real
and synthesized images which have had their spatial resolution artificially decreased by
bilinear interpolation 
(Figure \ref{fig:inception_score_plot}, top panels).
Note that as the spatial resolution is 
decreased, the accuracy decreases - indicating that resulting images contain less
class information (Figure \ref{fig:inception_score_plot}, scores below top panels).
We summarized this finding across all 1000 ImageNet classes for the ImageNet training
data (black),
a $128\times 128$ resolution AC-GAN (red) and a $64\times 64$ resolution AC-GAN (blue) in Figure \ref{fig:inception_score_plot} (bottom, left).
The black curve (clipped) provides an upper-bound on the discriminability of real images.

The goal of this analysis is to
show that synthesizing higher resolution images leads to increased discriminability.
The $128\times 128$ model achieves an accuracy of
10.1\% $\pm$ 2.0\% versus
7.0\% $\pm$ 2.0\% with samples resized to $64\times 64$
and
5.0\% $\pm$ 2.0\% with samples resized to $32\times 32$.
In other words, downsizing the outputs of the AC-GAN to $32 \times 32$ and $64 \times 64$ decreases
visual discriminability by 50\% and 38\% respectively.
Furthermore, 84.4\% of the ImageNet classes have higher accuracy at
$128\times 128$ than at $32 \times 32$ (Figure
\ref{fig:inception_score_plot}, bottom left).

We performed the same analysis on an AC-GAN trained to $64 \times 64$ spatial resolution.
This model achieved
less discriminability than a $128\times 128$ AC-GAN model.
Accuracies from the $64 \times 64$ model plateau at
a $64 \times 64$ spatial resolution consistent with previous results.
Finally, the $64\times 64$ resolution model achieves less
discriminability at 64 spatial resolution than the $128\times 128$ model.

To the best of our knowledge, this work is the first that attempts to measure
the extent to which an image synthesis model is `making use of its given output resolution',
and in fact is the first work to consider the issue at all.
We consider this an important contribution,
on par with proposing a model that synthesizes images from all 1000 ImageNet classes.
We note that the proposed method can be applied to any image synthesis model for which
a measure of `sample quality' can be constructed.
In fact, this method (broadly defined) can be applied to any type of synthesis model,
as long as there is an easily computable notion of sample quality and some
method for `reducing resolution'.  
In particular, we expect that a similar procecure can be carried out for audio synthesis.

\subsection{Measuring the Diversity of Generated Images} \label{section:diversity}

An image synthesis model is not very interesting if it only outputs one image.
Indeed, a well-known failure mode of GANs is that the generator will collapse and
output a single prototype that maximally fools the discriminator \citep{GANS, IMPROVEDTECHNIQUES}.
A class-conditional model of images is not very interesting if it only
outputs one image per class.
The Inception accuracy can not measure whether a model has collapsed.
A model that simply memorized one example from each
ImageNet class would do very well by this metric.
Thus, we seek a complementary metric to explicitly evaluate the intra-class perceptual
diversity of samples generated by the AC-GAN.

Several methods exist for quantitatively evaluating image similarity
by attempting to predict human perceptual similarity judgements.
The most successful 
of these is multi-scale structural similarity (MS-SSIM) \citep{MS-SSIM, MAD}.
MS-SSIM is a multi-scale variant of a well-characterized perceptual similarity metric
that attempts to discount aspects of an image that are not important for human
perception \citep{SSIM}.
MS-SSIM values range between 0.0 and 1.0; higher MS-SSIM values correspond to perceptually more similar
images.

\begin{figure}[t]
\captionsetup[subfigure]{labelformat=empty}
\setlength\tabcolsep{1.5pt}
\raisebox{-0.5\height}{\includegraphics[width=0.47\textwidth]{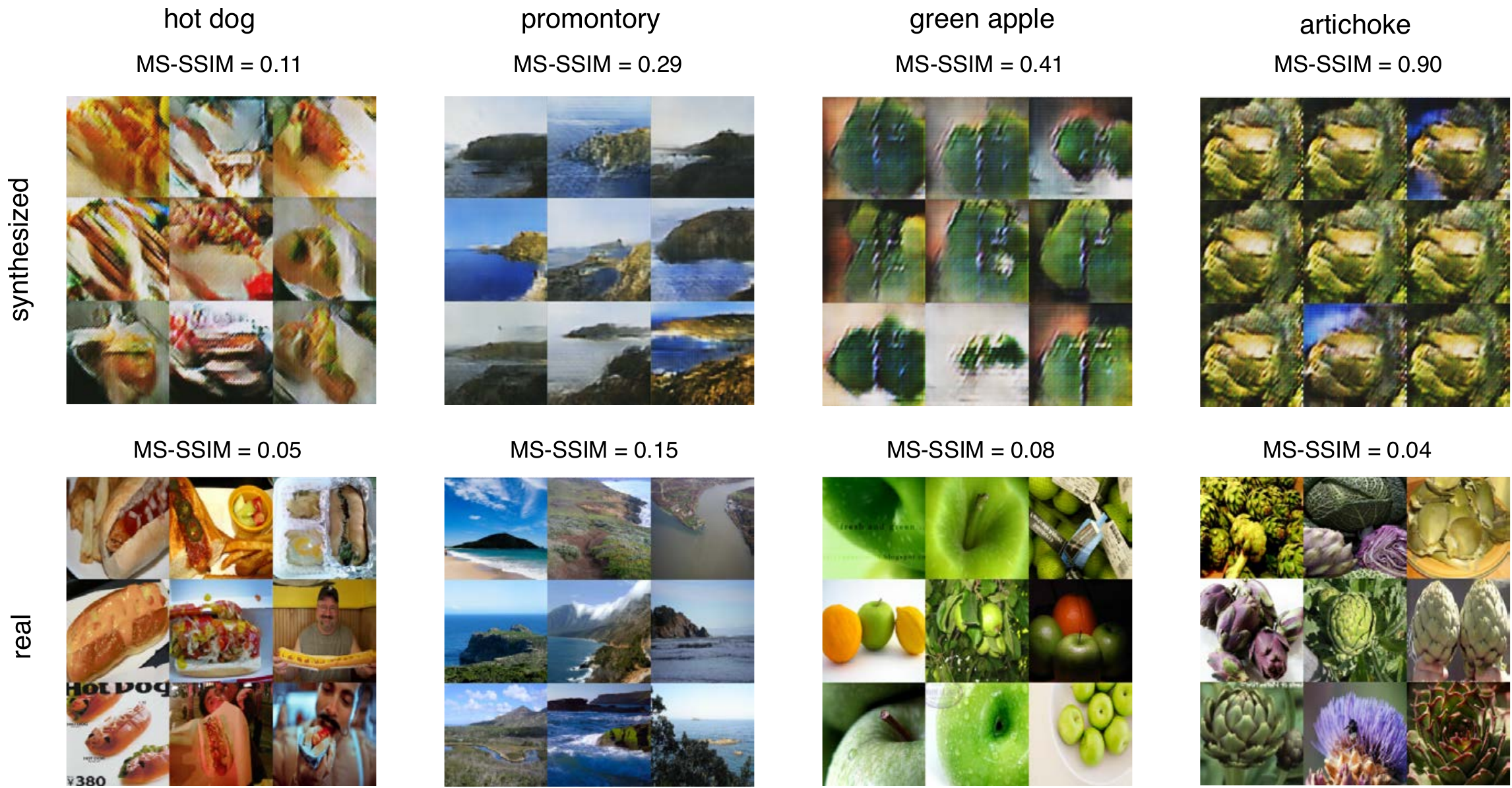}}

\centering
\caption{Examples of different MS-SSIM scores.
The top and bottom rows contain AC-GAN samples and training data, respectively.
}
\label{fig:ssimexamples}
\end{figure}

\begin{figure}[t]
\raisebox{-0.5\height}{\includegraphics[width=0.47\textwidth]{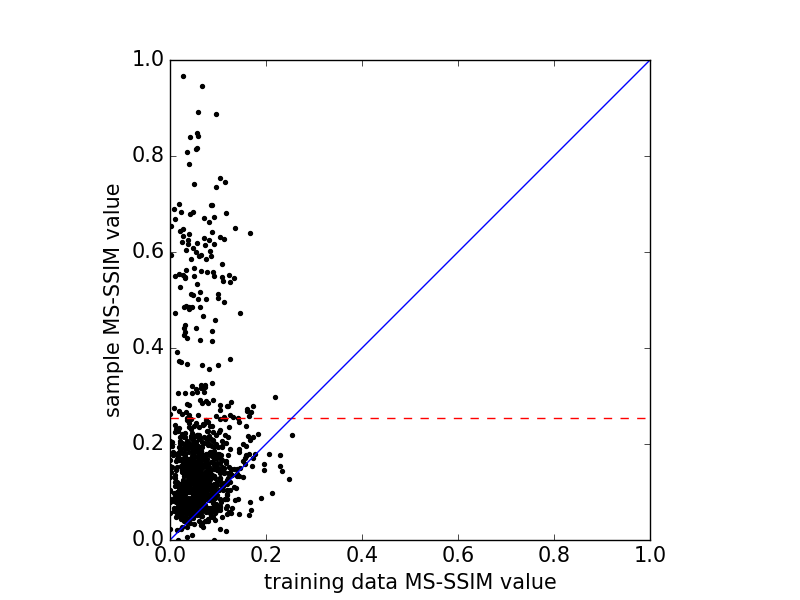}}
\centering
\caption{Comparison of the mean MS-SSIM scores between pairs of images within a given class for ImageNet training data and samples from the GAN (blue line is equality).
The horizontal red line marks the maximum MS-SSIM value (for training data) across all
ImageNet classes.
  Each point is an individual class.
  The mean score across the training data and the samples was 0.05 and 0.18 respectively.
  The mean standard deviation of scores across the training data and the samples was 0.06 and 0.08 respectively.
Scores below the red line (84.7\% of classes) arise from classes where GAN training largely
succeeded.
}
\label{fig:ssimhist}
\end{figure}

As a proxy for image diversity, we measure the MS-SSIM scores between 100 randomly chosen pairs of images within a given class.
Samples from classes that have higher diversity result in lower mean MS-SSIM scores (Figure \ref{fig:ssimexamples}, left columns); samples from classes with lower diversity have
higher mean MS-SSIM scores (Figure \ref{fig:ssimexamples}, right columns).
Training images from the ImageNet training data contain a variety of mean MS-SSIM scores across the classes indicating the variability of image diversity in ImageNet
classes (Figure \ref{fig:ssimhist}, x-axis).
Note that the highest mean MS-SSIM score (indicating the least variability) is 0.25 for the training data.

\begin{figure}[t]
\raisebox{-0.5\height}{

\includegraphics[width=0.47\textwidth]{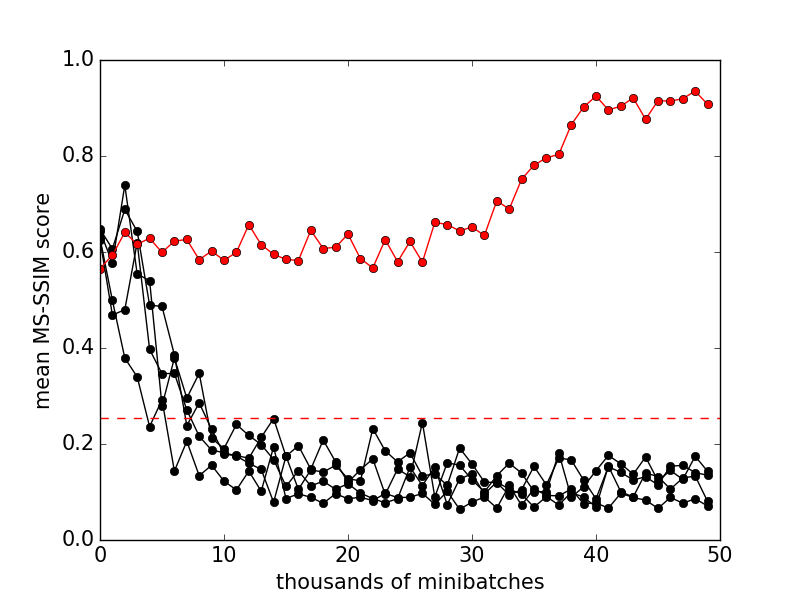}

}
\centering
\caption{Intra-class MS-SSIM for selected ImageNet classes throughout a training run.
Classes that successfully train (black lines) tend to have decreasing mean MS-SSIM scores.
Classes for which the generator `collapses' (red line) will have increasing mean MS-SSIM scores.
}
\label{fig:ssimhist2}
\end{figure}

We calculate the mean MS-SSIM score for all 1000 ImageNet classes generated by the
AC-GAN model.
We track this value during training to identify whether the generator has collapsed (Figure~\ref{fig:ssimhist2}, red curve).
We also employ this metric to compare the diversity of the training images to
the samples from the GAN model after training has completed.
Figure~\ref{fig:ssimhist} plots the mean MS-SSIM values for image samples and training
data broken up by class.
The blue line is the line of equality.
Out of the 1000 classes, we find that 847 have mean sample MS-SSIM scores below
that of the maximum MS-SSIM for the training data.
In other words, 84.7\% of classes have sample variability that exceeds
that of the least variable class from the ImageNet training data.

There are two points related to the MS-SSIM metric and our use of it that
merit extra attention.
The first point is that we are `abusing' the metric: 
it was originally intended to be used for measuring the quality of image
compression algorithms using a reference `original image'.
We instead use it on two potentially unrelated images.
We believe that this is acceptable for the following reasons:
First: visual inspection seems to indicate that the metric makes sense - pairs with higher MS-SSIM
do seem more similar than pairs with lower MS-SSIM.
Second: we restrict comparisons to images synthesized using the same class label.
This restricts use of MS-SSIM to situations more similar to those in which it
is typically used (it is not important which image is the reference).
Third: the metric is not `saturated' for our use-case. If most scores were around 0,
then we would be more concerned about the applicability of MS-SSIM.
Finally: The fact that training data achieves more variability by this metric (as expected)
is itself evidence that the metric is working as intended.

The second point is that the MS-SSIM metric is not intended as a proxy
for the entropy of the generator distribution in pixel space,
but as a measure of perceptual diversity of the outputs.
The entropy of the generator output distribution is hard to compute
and pairwise MS-SSIM scores would not be a good proxy.
Even if it were easy to compute, we argue that it would still
be useful to have a separate measure of perceptual diversity.
To see why, consider that the generator entropy 
will be sensitive to trivial changes in contrast
as well as changes in the semantic content of the outputs.
In many applications, we don't care about this contribution to the entropy,
and it is useful to consider measures that attempt to ignore
changes to an image that we consider `perceptually meaningless',
hence the use of MS-SSIM.

\subsection{Generated Images are both Diverse and Discriminable} \label{section:correlation}

\begin{figure}[h]
\centering\includegraphics[width=0.49\textwidth]{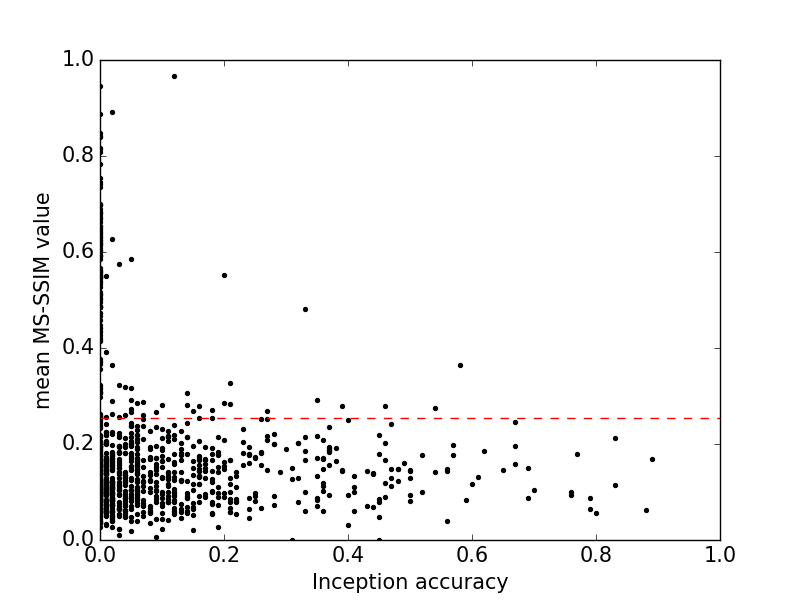}%

\caption{
Inception accuracy vs MS-SSIM for all 1000 ImageNet classes ($r^2 = -0.16$).
Each data point represents the mean MS-SSIM value for samples from one class.
As in Figure \ref{fig:ssimhist}, the red line marks the maximum MS-SSIM value (for training data)
across all ImageNet classes.
Samples from AC-GAN models do not achieve variability at the expense of discriminability.
}\vspace{-10pt}
\label{fig:ssimvinceptionfig}
\end{figure}

We have presented quantitative metrics demonstrating that AC-GAN
samples may be diverse and discriminable but we have yet
to examine how these metrics interact.
Figure \ref{fig:ssimvinceptionfig} shows the joint distribution of Inception accuracies
and MS-SSIM scores across all classes.
Inception accuracy and MS-SSIM are anti-correlated ($r^2 = -0.16$).
In fact, 74\% of the classes with low diversity (MS-SSIM $\geq 0.25$)
contain Inception accuracies $\leq 1\%$.
Conversely, 78\% of classes with high diversity (MS-SSIM $< 0.25$)
have Inception accuracies that exceed 1\%. In comparison, the Inception-v3 model
achieves 78.8\% accuracy on average across all 1000 classes \citep{szegedy2015rethinking}.
A fraction of
the classes AC-GAN samples reach this level of accuracy.
This indicates opportunity for future image synthesis
models. 

These results suggest that GANs that drop modes are most likely to produce
low quality images.
This stands in contrast to a popular hypothesis about GANs,
which is that they achieve high sample quality at the expense of variability.
We hope that these findings can help structure further investigation
into the reasons for differing sample quality between GANs and
other image synthesis models.

\subsection{Comparison to Previous Results} \label{section:inceptionscore}

Previous quantitative results for image synthesis models trained on ImageNet
are reported in terms of log-likelihood \citep{PIXELRNN, PIXELCNN}.
Log-likelihood is a coarse and potentially inaccurate measure of sample quality
\citep{TheisEtAl}. 
Instead we compare with previous state-of-the-art results on CIFAR-10
using a lower spatial resolution ($32\times 32$).
Following the procedure in \cite{IMPROVEDTECHNIQUES},
we compute the Inception score\footnote{
The Inception score is given by
$\exp\left( E_x [D_{\small KL}(p(y|x) \; || \; p(y))] \right)$
where $x$ is a particular image, $p(y|x)$ is the conditional output distribution over
the classes in a pre-trained Inception network \citep{INCEPTION} given $x$,
and $p(y)$ is the marginal distribution over the classes.
}
for 50000 samples from an AC-GAN with resolution ($32\times 32$),
split into 10 groups at random.
We also compute the Inception score for 25000 extra samples,
split into 5 groups at random.
We select the best model based on the first score and report the second score.
Performing a grid search across 27 hyperparameter configurations,
we are able to achieve a score of 8.25 $\pm$ 0.07
compared to state of the art 
8.09 $\pm$ 0.07 \citep{IMPROVEDTECHNIQUES}.
Moreover, we accomplish this without employing any of the new techniques
introduced in that work (i.e. virtual batch normalization,
minibatch discrimination, and label smoothing).

\begin{figure}[!h]
\includegraphics[width=0.47\textwidth]{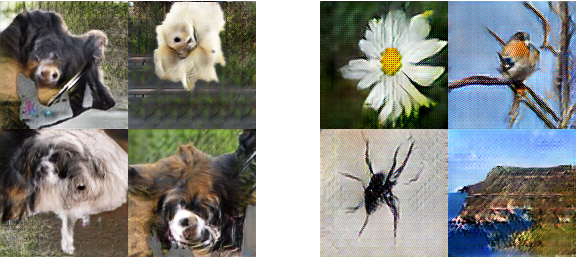}
\centering
\caption{
  Samples generated from the ImageNet dataset.
  (Left) Samples generated from the model in \cite{IMPROVEDTECHNIQUES}.
  (Right) Randomly chosen samples generated from an AC-GAN.
  AC-GAN samples possess global coherence absent from the samples of the earlier model.
}
\label{fig:qualitative}
\end{figure}

This provides additional evidence that AC-GANs are effective even without
the benefit of class splitting.
See Figure \ref{fig:qualitative} for a qualitative comparison
of samples from an AC-GAN and samples from the model in \cite{IMPROVEDTECHNIQUES}.

\subsection{Searching for Signatures of Overfitting}
\begin{figure}[!h]
\includegraphics[width=0.47\textwidth]{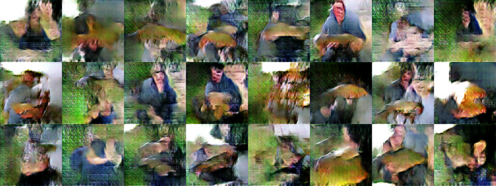}
\includegraphics[width=0.47\textwidth]{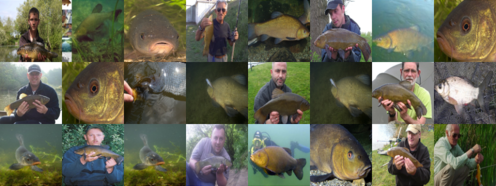}
\centering
\caption{Nearest neighbor analysis. (Top) Samples from a single ImageNet class. (Bottom)
Corresponding nearest neighbor (L1 distance) in training data for each sample.}
\label{fig:neighbors}
\end{figure}

One possibility that must be investigated is that
the AC-GAN has overfit on the training data.
As a first check that the network does not memorize the
training data, we identify the nearest neighbors of image samples in
the training data measured by L1 distance in pixel space (Figure \ref{fig:neighbors}).
The nearest neighbors from the training data do not resemble the corresponding samples.
This provides evidence that the AC-GAN is not merely memorizing the training data.

\begin{figure}[!h]
\includegraphics[width=0.47\textwidth]{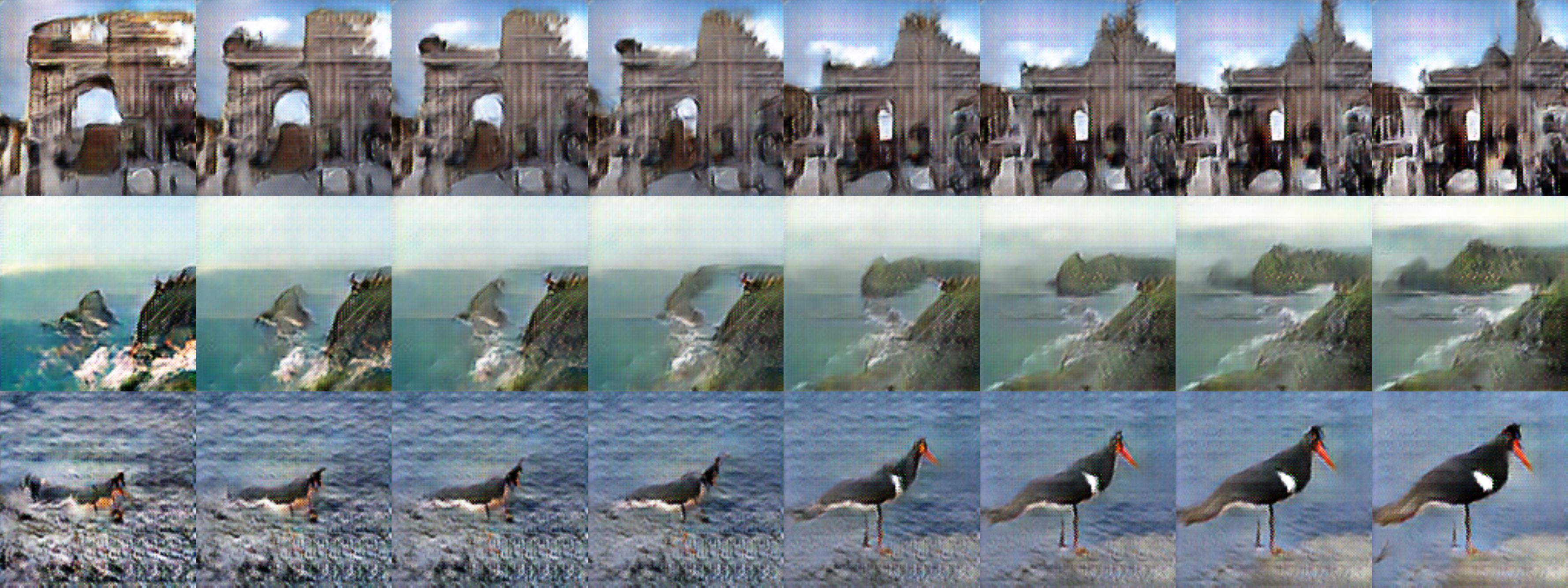}
\includegraphics[width=0.47\textwidth]{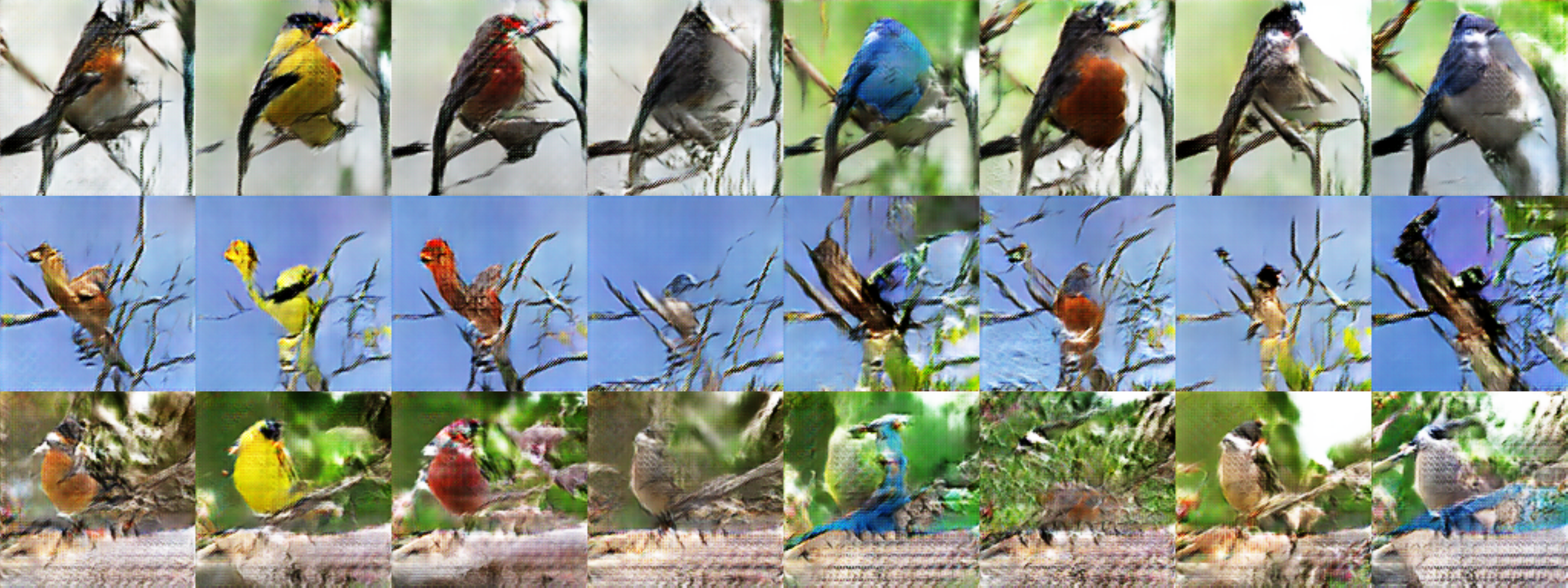}
\centering
\caption{(Top) Latent space interpolations for selected ImageNet classes.
Left-most and right-columns show three pairs of image samples - each pair from a distinct class.
Intermediate columns highlight linear interpolations in the latent space between these three
pairs of images.
(Bottom) Class-independent information contains global structure about the synthesized image.
Each column is a distinct bird class while each row corresponds to a fixed latent code $z$.
}
\label{fig:interpolations}
\end{figure}

A more sophisticated method for understanding the degree of overfitting in
a model is to explore that model's latent space by interpolation.
In an overfit model one might observe discrete transitions
in the interpolated images and regions in
latent space that do not correspond to meaningful images \citep{INTERPOLATIONS, DCGAN, nvp}.
Figure \ref{fig:interpolations} (Top) highlights interpolations in the latent space between several image samples.
Notably, the generator learned that certain combinations of dimensions
correspond to semantically meaningful features (e.g. size of the arch, length of a
bird's beak) and there are no discrete transitions or `holes' in the
latent space.

A second method for exploring 
the latent space of the AC-GAN is to exploit the structure of the model.
The AC-GAN factorizes its representation into class information
and a class-independent latent representation $z$.
Sampling the AC-GAN with $z$ fixed but altering the class label
corresponds to 
generating samples with the same `style' across multiple classes \citep{SSVAE}.
Figure \ref{fig:interpolations} (Bottom) shows samples from 8 bird classes.
Elements of the same row have the same $z$.
Although the class changes for each column, elements
of the global structure (e.g. position, layout, background) are preserved,
indicating that AC-GAN can represent certain types of `compositionality'. 

\subsection{Measuring the Effect of Class Splits on Image Sample Quality.} \label{section:count}

\begin{figure}[!h]
\includegraphics[width=0.50\textwidth]{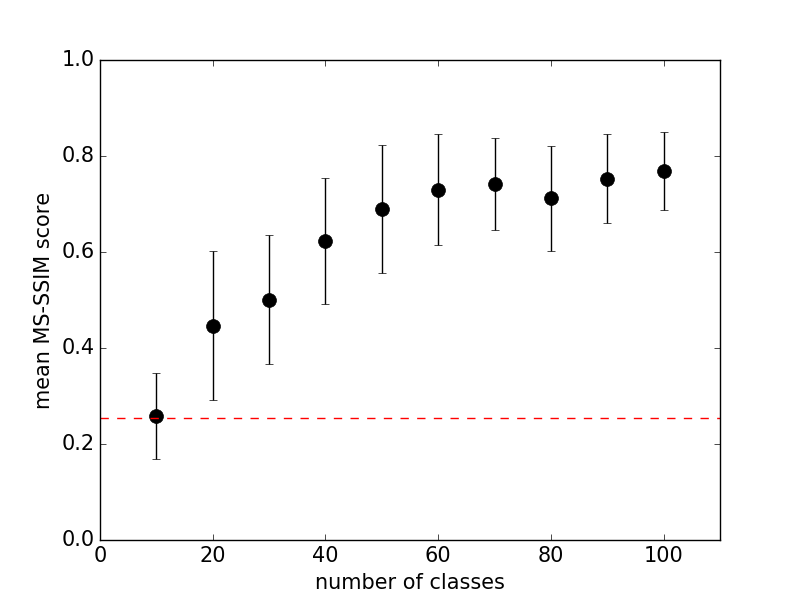}
\centering
\caption{Mean pairwise MS-SSIM values for 10 ImageNet classes plotted
  against the number of ImageNet classes used during training.
  We fix everything except the number of classes trained on,
  using values from 10 to 100.
  We only report the MS-SSIM values for the first 10 classes
  to keep the scores comparable.
  MS-SSIM quickly goes above 0.25 (the red line) as the class count
  increases.
  These scores were computed using 9 random restarts per class count,
  using the same number of training steps for each model.
  Since we have observed that generators do not recover from the collapse phase,
  the use of a fixed number of training steps seems justified in this case.
}
\label{fig:classcount}
\end{figure}

Class conditional image synthesis affords the opportunity to divide up
a dataset based on image label.
In our final model we divide 1000 ImageNet classes across 100 AC-GAN models. In this section
we describe experiments that highlight the benefit of cutting down the diversity of classes for
training an AC-GAN.
We employed an ordering of the labels and divided it into contiguous groups of 10.
This ordering can be seen in the following section, where we display
samples from all 1000 classes.
Two aspects of the split merit discussion: the number of classes per split and
the intra-split diversity.
We find that training a fixed model on more classes harms
the model's ability to produce compelling samples
(Figure \ref{fig:classcount}).
Performance on larger splits can
be improved by giving the model more parameters.
However, using a small split is not sufficient to achieve good performance.
We were unable to train a GAN \citep{GANS} to converge reliably even for a split size of 1.
This raises the question of whether it is easier to train a model on a
diverse set of classes than on a similar set of classes:
We were unable to find conclusive evidence that the selection
of classes in a split significantly affects sample quality.

We don't have a hypothesis about what causes this sensitivity to class count that is well-supported experimentally.
We can only note that, since the failure case that occurs when the class count is increased is `generator collapse',
it seems plausible that general methods for addressing `generator collapse' could also address this sensitivity.

\subsection{Samples from all 1000 ImageNet Classes}

We also generate 10 samples from each of the 1000 ImageNet classes, hosted
\href{https://goo.gl/photos/8bgHBkCwDEVTXAPaA}{here}.
As far as we are aware, no other image synthesis work has included a similar analysis.

\section{Discussion}

This work introduced the AC-GAN architecture and demonstrated that AC-GANs
can generate globally coherent ImageNet samples.
We provided a new quantitative metric for image discriminability as a
function of spatial resolution.
Using this metric we demonstrated that our samples are more discriminable than
 those from a model
that generates lower resolution images and performs a naive resize operation.
We also analyzed the diversity of our samples with respect to the training data and 
provided some evidence that the image samples from the majority of classes are
comparable in diversity to ImageNet data.

Several directions exist for building upon this work.
Much work needs to be done to improve the visual discriminability
of the $128\times 128$ resolution model.
Although some synthesized image classes exhibit high Inception accuracies, 
the average Inception accuracy of the model
($10.1\% \pm 2.0\%$) is still far below real training data at 81\%.
One immediate opportunity for addressing this is to
augment the discriminator with a pre-trained
model to perform additional supervised tasks (e.g. image segmentation, \cite{UNET}).

Improving the reliability of GAN training is an ongoing research topic.
Only 84.7\% of the ImageNet classes exhibited diversity
comparable to real training data.
Training stability was vastly aided by dividing up 1000 ImageNet classes across
100 AC-GAN models. Building a single 
model that could generate samples from all 1000 classes would be 
an important step forward.

Image synthesis models provide a unique opportunity for performing semi-supervised
learning: these models build a rich prior over natural image statistics
that can be leveraged by classifiers to improve predictions on datasets for which
few labels exist.
The AC-GAN model can perform semi-supervised learning
by ignoring the component of the loss arising from class labels when a label is
unavailable for a given training image.
Interestingly, prior work suggests that achieving good sample quality
might be independent of success in semi-supervised learning
\citep{IMPROVEDTECHNIQUES}.

\newpage

\newpage

\bibliography{example_paper}
\bibliographystyle{icml2017}

\newpage

\appendix
\onecolumn
\section{Hyperparameters}

We summarize hyperparameters used for the ImageNet model in Table \ref{tab:model_description} and
for the CIFAR-10 model in Table \ref{tab:cifar_model_description}.

\begin{table*}[h]
\centering
\resizebox{\linewidth}{!}{
\begin{tabular}{@{}rllllll@{}} \toprule
Operation              & Kernel       & Strides      & Feature maps & BN?          & Dropout & Nonlinearity \\ \midrule
$G_x(z)$ -- $110 \times 1 \times 1$ input                                                                 \\
Linear                 & N/A          & N/A          & $768$        & $\times$     & 0.0     & ReLU       \\
Transposed Convolution & $5 \times 5$ & $2 \times 2$ & $384$        & $\surd$      & 0.0     & ReLU       \\
Transposed Convolution & $5 \times 5$ & $2 \times 2$ & $256$        & $\surd$      & 0.0     & ReLU       \\
Transposed Convolution & $5 \times 5$ & $2 \times 2$ & $192$        & $\surd$      & 0.0     & ReLU       \\
Transposed Convolution & $5 \times 5$ & $2 \times 2$ & $3$          & $\times$     & 0.0     & Tanh       \\
$D(x)$ -- $128 \times 128 \times 3$ input                                                                   \\
Convolution            & $3 \times 3$ & $2 \times 2$ & $16$         & $\times$     & 0.5     & Leaky ReLU \\
Convolution            & $3 \times 3$ & $1 \times 1$ & $32$         & $\surd$      & 0.5     & Leaky ReLU \\
Convolution            & $3 \times 3$ & $2 \times 2$ & $64$         & $\surd$      & 0.5     & Leaky ReLU \\
Convolution            & $3 \times 3$ & $1 \times 1$ & $128$        & $\surd$      & 0.5     & Leaky ReLU \\
Convolution            & $3 \times 3$ & $2 \times 2$ & $256$        & $\surd$      & 0.5     & Leaky ReLU \\
Convolution            & $3 \times 3$ & $1 \times 1$ & $512$        & $\surd$      & 0.5     & Leaky ReLU \\
Linear                 & N/A          & N/A          & $11$         & $\times$     & 0.0     & Soft-Sigmoid\\
Optimizer              & \multicolumn{6}{@{}l@{}}{Adam ($\alpha = 0.0002$, $\beta_1 = 0.5$, $\beta_2 = 0.999$)}  \\
Batch size             & \multicolumn{6}{@{}l@{}}{100}												      \\
Iterations             & \multicolumn{6}{@{}l@{}}{50000}											      \\
Leaky ReLU slope       & \multicolumn{6}{@{}l@{}}{0.2}                                                   \\
Weight, bias initialization  & \multicolumn{6}{@{}l@{}}{Isotropic gaussian ($\mu = 0$, $\sigma = 0.02$), Constant($0$)} \\ \bottomrule
\end{tabular}}
\vspace{0.2cm}
\caption{\label{tab:model_description} ImageNet hyperparameters.
  A Soft-Sigmoid refers to an operation over $K+1$ output units where we apply a Softmax activation to $K$ of
  the units and a Sigmoid activation to the remaining unit.
We also use activation noise in the discriminator as suggested in \cite{IMPROVEDTECHNIQUES}.}
\end{table*}

\begin{table*}[h]
\centering
\resizebox{\linewidth}{!}{
\begin{tabular}{@{}rllllll@{}} \toprule
Operation               & Kernel       & Strides      & Feature maps & BN?          & Dropout & Nonlinearity \\ \midrule
$G_x(z)$ -- $110 \times 1 \times 1$ input                                                                 \\
Linear                  & N/A          & N/A          & $384$        & $\times$     & 0.0     & ReLU       \\
Transposed Convolution  & $5 \times 5$ & $2 \times 2$ & $192$        & $\surd$      & 0.0     & ReLU       \\
Transposed Convolution  & $5 \times 5$ & $2 \times 2$ & $96$         & $\surd$      & 0.0     & ReLU       \\
Transposed Convolution  & $5 \times 5$ & $2 \times 2$ & $3$          & $\times$     & 0.0     & Tanh       \\
$D(x)$ -- $32 \times 32 \times 3$ input                                                                   \\
Convolution             & $3 \times 3$ & $2 \times 2$ & $16$         & $\times$     & 0.5     & Leaky ReLU \\
Convolution             & $3 \times 3$ & $1 \times 1$ & $32$         & $\surd$      & 0.5     & Leaky ReLU \\
Convolution             & $3 \times 3$ & $2 \times 2$ & $64$         & $\surd$      & 0.5     & Leaky ReLU \\
Convolution             & $3 \times 3$ & $1 \times 1$ & $128$        & $\surd$      & 0.5     & Leaky ReLU \\
Convolution             & $3 \times 3$ & $2 \times 2$ & $256$        & $\surd$      & 0.5     & Leaky ReLU \\
Convolution             & $3 \times 3$ & $1 \times 1$ & $512$        & $\surd$      & 0.5     & Leaky ReLU \\
Linear                  & N/A          & N/A          & $11$         & $\times$     & 0.0     & Soft-Sigmoid\\
Generator Optimizer     & \multicolumn{6}{@{}l@{}}{Adam ($\alpha = [0.0001, 0.0002, 0.0003]$, $\beta_1 = 0.5$, $\beta_2 = 0.999$)}  \\
Discriminator Optimizer & \multicolumn{6}{@{}l@{}}{Adam ($\alpha = [0.0001, 0.0002, 0.0003]$, $\beta_1 = 0.5$, $\beta_2 = 0.999$)}  \\
Batch size              & \multicolumn{6}{@{}l@{}}{100}												      \\
Iterations              & \multicolumn{6}{@{}l@{}}{50000}											      \\
Leaky ReLU slope        & \multicolumn{6}{@{}l@{}}{0.2}                                                   \\
Activation noise standard deviation & \multicolumn{6}{@{}l@{}}{$[0,0.1,0.2]$}                                                   \\
Weight, bias initialization  & \multicolumn{6}{@{}l@{}}{Isotropic gaussian ($\mu = 0$, $\sigma = 0.02$), Constant($0$)} \\ \bottomrule
\end{tabular}}
\vspace{0.2cm}
\caption{\label{tab:cifar_model_description} CIFAR-10 hyperparameters.
  When a list is given for a hyperparameter it means that we performed a grid search using the values in the list.
  For each set of hyperparameters, a single AC-GAN was trained on the whole CIFAR-10 dataset.
  For each AC-GAN that was trained, we split up the samples into groups so that we could give some sense of the variance of the Inception Score.
  To the best of our knowledge, this is identical to the analysis performed in \cite{IMPROVEDTECHNIQUES}.
}
\end{table*}

\end{document}